\documentclass[10pt,twocolumn]{article}
\UseRawInputEncoding
\usepackage{lmodern, upquote}
\usepackage{geometry}
\usepackage{titlesec}

\titlespacing{\section}{0em}{*1}{*1}[0em]
\titlespacing{\subsection}{0em}{*1}{*0.7}[0em]
\titlespacing{\subsubsection}{0em}{*0.8}{*0.5}[0em]
\titlespacing{\paragraph}{0em}{*0.5}{*0.5}[0em]

\newenvironment{resume}{\thispagestyle{empty}\subsection*{Résumé}\em}{}

\renewenvironment{abstract}{\subsection*{Abstract}\em}{}
\newenvironment{keywords}{\subsection*{Keywords}\em}{}

\usepackage{amsmath, amsfonts, amssymb}
\usepackage{graphicx}
\usepackage[hyphens]{url}
\usepackage{hyperref}
\hypersetup{colorlinks=true,citecolor=blue}
\usepackage[table,svgnames,x11names,dvipsnames]{xcolor}
\usepackage{multirow, siunitx}
\usepackage{placeins} 
\usepackage{cleveref}

\title{A Benchmarking on Cloud based Speech-To-Text Services for French Speech and Background Noise Effect\thanks{6th National Conference on Practical Applications of Artificial Intelligence, 2021, Bordeaux, France}} 
\author{Binbin Xu\textsuperscript{1*}, Chongyang Tao\textsuperscript{1+}, Zidu Feng\textsuperscript{1+}, Youssef Raqui\textsuperscript{2}, Sylvie Ranwez\textsuperscript{1*} 
\\[6pt]
\textsuperscript{1} EuroMov Digital Health in Motion, Univ Montpellier, IMT Mines Alès\\
\textsuperscript{2} DiappyMed}
\date{\texttt{\textsuperscript{1*}firstname.lastname@mines-ales.fr \textsuperscript{1+}firstname.lastname@mines-ales.org}}

\begin{document}
\maketitle

\begin{resume}
{Alors que les applications de reconnaissance vocale se sont imposées dans notre quotidien, il existe peu d'études à grande échelle pour comparer les performances des solutions de l'état de l'art. Ceci est d'autant plus vrai dans une langue autre que la langue anglaise. Cet article propose une telle analyse comparative basée sur 17 heures d'enregistrement en Français. Quatre systèmes sont analysés : {Google Cloud Speech-To-Text},  {Microsoft Azure Cognitive Services}, {Amazon Transcribe}, et {IBM Watson Speech to Text}. Chacun ayant été mis à l'épreuve de cinq niveaux de bruit de fond, c'est l'équivalent de 400 heures de discours qui sont analysées. {Microsoft Azure Cognitive Services} a montré les meilleurs résultats en terme de taux d'erreur et une bonne résistance au bruit, tandis que la sensibilité au bruit d'{IBM Watson Speech to Text} compromet son usage en situation réelle.   }
\end{resume}

\begin{abstract}
This study presents a large scale benchmarking on cloud based Speech-To-Text systems: {Google Cloud Speech-To-Text},  {Microsoft Azure Cognitive Services}, {Amazon Transcribe}, {IBM Watson Speech to Text}. 
For each systems, $40\,158$ clean and noisy speech files about $101$ hours are tested. 
Effect of background noise on STT quality is also evaluated with 5 different Signal-to-noise ratios from \SI{40}{\decibel} to \SI{0}{\decibel}.
Results showed that {Microsoft Azure} provided lowest transcription error rate $9.09\%$ on clean speech, with high robustness to noisy environment. 
{Google Cloud} and {Amazon Transcribe} gave similar performance, but the latter is very limited for time-constraint usage. 
Though {IBM Watson} could work correctly in quiet conditions, it is highly sensible to noisy speech which could strongly limit its application in real life situations. 
\end{abstract}


\section{Introduction}

Lots applications with automated speech recognition (ASR) or Speech-To-Text (STT) over the past few years have been developed to improve our daily life like personal voice assistant, or have been deeply integrated in many of business chains. 
Thanks to the substantial development of deep neural network (DNN), the performance of STT has been drastically improved. 
Like other deep neural network applications, today it is not surprising that in some situations, current STT can even outperform humans. 
The IBM/Appen human transcription study \cite{Saon2017English} showed that word error rate of human parity is about $5.1\%$. Microsoft Research is the first team reaching this milestone. 
However, the outstanding performances in DNN is based on large amount of labeled training data. This is also the case for DNN models on STT. 
For languages other than English, there's much less high quality audio data like in English. In consequence, the performances of STT on other languages are in general lower than for English, especially for languages featuring rich morphology like French. 

Though many public Deep Neural Network models are available for offline use, retraining or regular updating require extensive computing power which prevents individuals or small business from accessing these models or using them in an efficient way. 
The choice will be the cloud-based API services. 
Actually, the most powerful STT systems are all cloud-based. Integrating these systems in an application or a product line requires at first a benchmarking on their performance. 
There exist many benchmarking studies on the performance of cloud-based STT services. However, they are often conducted with very small or small sample size, for example, 20--60 sentences or hundreds of sentences. The benchmarking on English from Picovoice is one of the few large scale tests on STT, which contains 2620 audio files (5h24m) from LibriSpeech dataset \cite{Picovoice2020Speech}.
Benchmarking of cloud-based STT on French is even less studied. 
Another major negative factor on STT performance is the background noise. Very often, only clean speech record is processed. However, for most real-life application, the background noise can hardly be avoided. This should be taken into account in STT benchmarking as well. 

The objective of this study is to benchmark four most used Speech-To-Text API (Application Programming Interface) with a large French dataset : 6693 files, about 17hours speech record. Five levels of common background noise are added in the clean speech and evaluated additionally. In total, more than 400 hours speech are transcribed.

\section{Speech-To-Text system and data}

\subsection{Cloud based STT services}

Four cloud based Speech-To-Text services are evaluated in this work:

\begin{itemize}

\item {Amazon Transcribe}, is part of Cloud Computing Services from the Amazon Web Services (AWS) which holds currently the largest share in Cloud Computing market. 
Their recent speech recognition model on English reached State-of-the-Art word error rate at $6.2\%$ \cite{Chiu2018State}. 
To convert speech files to text, the data needs to be at first uploaded to Amazon Simple Storage Service (Amazon S3). Then Transcribe call the objects from S3 for transcription. 
Though Transcribe jobs can be treated on batch mode (up to 100 parallel jobs). This S3 requirement adds additional complexity for the transcription tasks. Actually {Amazon Transcribe} is the only STT requiring storage. The other three services can be feed directly with audio files. 

\item {Google Cloud Speech-to-Text}, is integrated in the widely used platform Google Cloud. 
In 2012, Google Research had achieved word error rate at $15\%$ for English broadcast news transcription. 
This error rate dropped considerably to $5.6\%$ with updated model trained on over $12\,500$ hours audio in 2018 \cite{Chiu2018State}. 
Their STT model is one of the most powerful in the market, and the performance is continuously improving. 

\item {IBM Watson Speech-to-Text}. IBM Watson is a conventional top player in speech recognition. In 2015, their speech recognition system beat other models with a word error rate at $8\%$ \cite{Saon2015IBM}. Two years later, their system reached $5.5\%$ \cite{Saon2017English}. It's now among the most popular STT services and provides similar features as other cloud STT. 

\item {Microsoft Azure Cognitive Services}. Microsoft's speech recognition is now one of the leading STT service. In 2017, their model reached a historical human parity milestone on conversational telephony speech transcription, with 5.1\% word error rate in benchmarked Switchboard task \cite{Xiong2018Microsoft}. As all the other STT systems, Microsoft's STT system is also integrated in the Cloud Computing platform.

\end{itemize}

All the four STT services offer the possibility to customize (like domain-specific) or retrain the Speech-to-Text models. However, since they are all black-boxed APIs, the background models and architectures are unknown, it is difficult to benchmark the customized models with different configurations in a fair way. So, only the basic models (APIs) are called.

\subsection{Speech corpus}

The basic audio dataset in this work is from WCE-SLT-LIG \cite{Besacier2014Word, Le2016Joint}. 
This corpus contains 6693 speech utterances recorded by 42 native speakers. 
\begin{figure}[!htb]
\centering
\includegraphics[width=0.95\columnwidth]{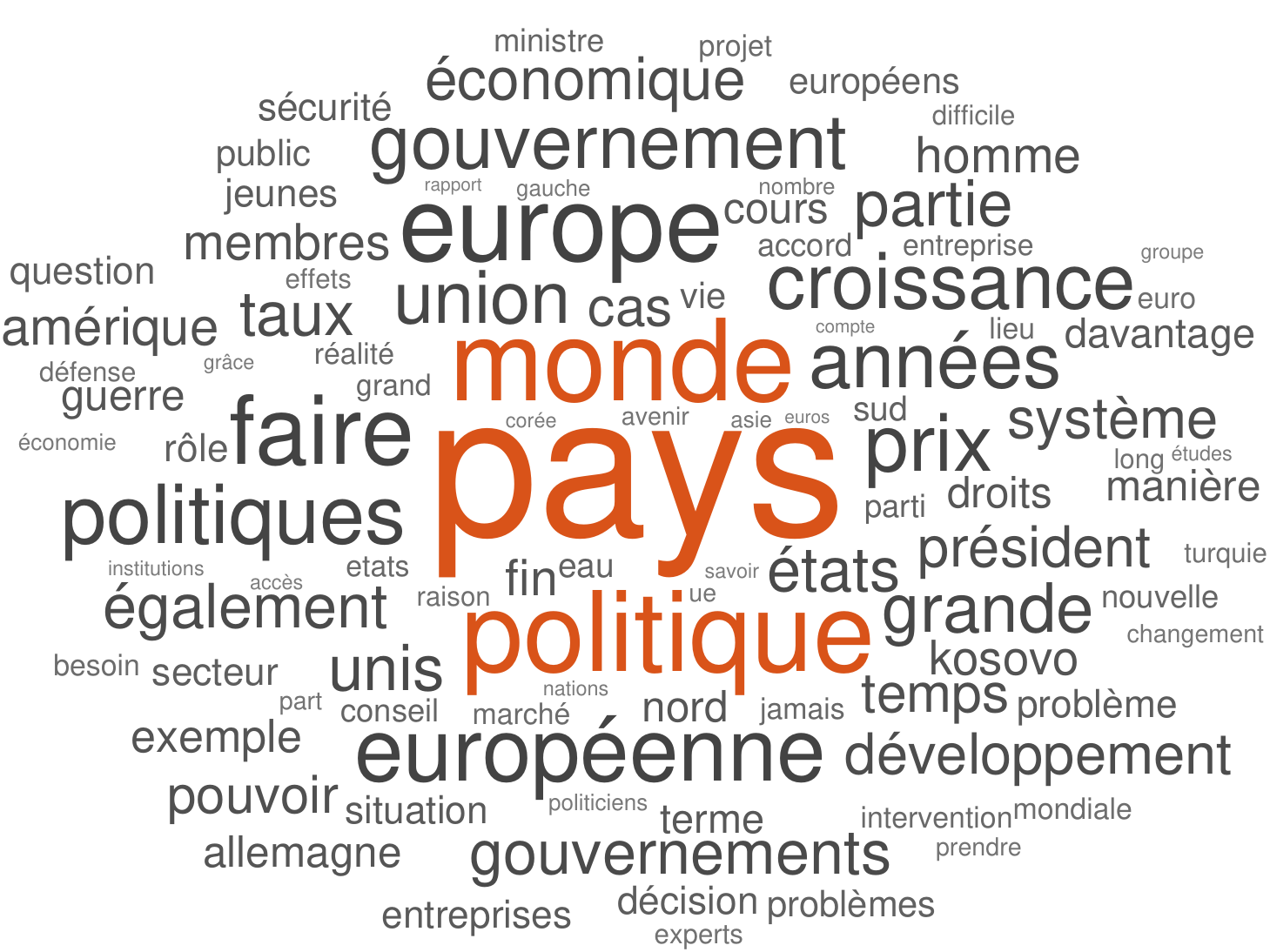}
\caption{Main topic of WCE-SLT-LIG corpus}
\label{fig:cleanaudio_wordcloud}%
\end{figure}
\FloatBarrier
They come from French news media with main topic on European economy. The total audio duration is 16h52. The ground-truth transcriptions are also available, which makes the benchmarking possible. 

\begin{figure}[!htb]
\centering
\includegraphics[width=0.95\columnwidth]{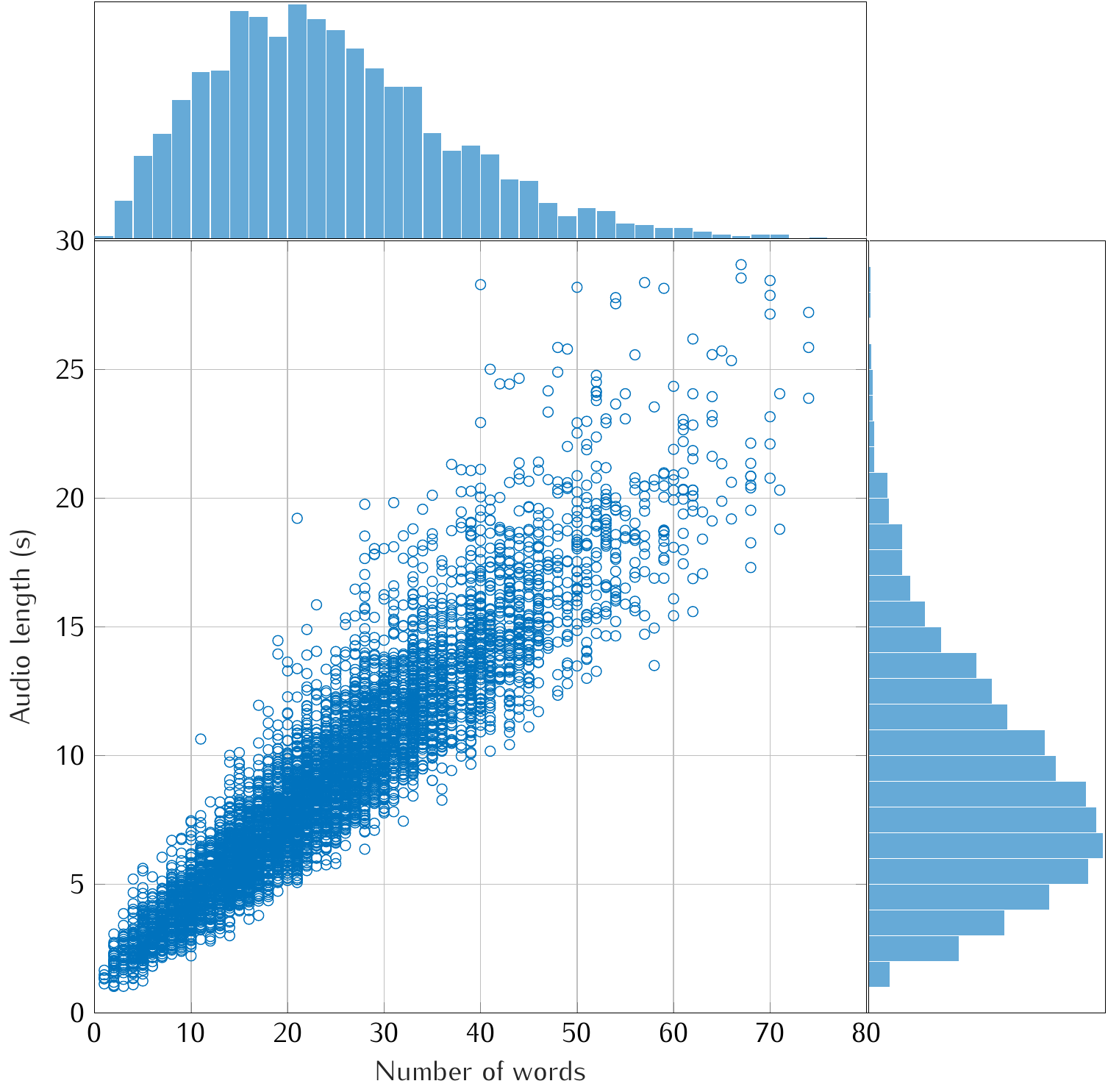}
\caption{Distribution of WCE-SLT-LIG corpus}
\label{fig:cleanaudio_hist}
\end{figure}

The number of word in this corpus is $22 \pm 12.8$ (median $\pm$ standard deviation), with audio duration $8.4 \pm 4.6$ seconds as shown in \Cref{fig:cleanaudio_hist}. 

\subsection{Environmental noise corpus}

In real world cases, most speech takes place in noisy environments. This is one of the main challenges in Speech-to-Text applications. 
To evaluate the effects on the STT quality, we introduce another recently released environmental noise dataset: Microsoft Scalable Noisy Speech Dataset (MS-SNSD) \cite{Reddy2019Scalable}. 
The dataset provides a variety of common environmental noise, which can be mixed on clean speech data. The signal-to-noise (SNR) in \SI{}{\decibel} can be configured as well. 
\begin{equation*}
{\displaystyle \mathrm {SNR_{dB}} =10\log _{10}\left({\frac {P_{\mathrm {signal} }}{P_{\mathrm {noise} }}}\right).}
\end{equation*}
where $P_{\mathrm {signal} }$, $P_{\mathrm {noise} }$ are the power of signal and background noise. 
We set 5 SNR cases here: \SI{40}{\decibel}, \SI{30}{\decibel}, \SI{20}{\decibel}, \SI{10}{\decibel} and \SI{0}{\decibel} (1:1 signal vs. noise). 

The raw MS-SNSD contains 181 noise files. However, many of them are recorded with strong conversations in other languages (English, German etc.). Some of noise type are also less common. So, these noises are excluded. We'd like to evaluate the effect of noise type on the performance of STT, to make sure that some types are not over-presented, 96 noise files in 18 types are kept. 

\begin{table}[!htbp]
\scriptsize
  \centering
    \begin{tabular}{>{\ttfamily}{l}>{\ttfamily}{l}>{\ttfamily}{l}} 
    AirConditioner & Kitchen & SqueakyChair \\
    AirportAnnouncements & LivingRoom & Station \\
    Babble & Munching & Traffic \\
    Cafe  & Restaurant & Typing \\
    CafeTeria & ShuttingDoor & VacuumCleaner \\
    CopyMachine & Square & WasherDryer \\
    \end{tabular}%
  \label{tab:noise}%
  \caption{Types of background noise used in this work. 96 noises in 18 types}
\end{table}%

\subsection{Evaluation metrics}

In Speech-to-Text, the most commonly used metric to evaluate the performance is {\sc Word error rate} (WER). Other metrics exist, like {\sc Match error rate} (MER); {\sc Word information lost} (WIL) or {\sc Word information preserve} (WIP) \cite{Morris2004From}. 
\begin{align}
   WER & = \frac{S + D + I}{N1 = H + S + D}\\
   MER & = \frac{S + D + I}{N = H + S + D + I}\\
   WIP & = \frac{H}{N_1}\cdot\frac{H}{N_2} \cong \frac{I(X,Y)}{H(Y)},\\
   WIL & = 1 - WIP
\end{align}
where $H$, $S$, $D$ and $I$ correspond to the total number of word hits, substitutions, deletions and insertions. 
$N_1$ and $N_2$ are respectively the number of words in ground-truth text and the output transcripts. 
The lower are WER, MER and WIL, the better the performance is.

\begin{table}[htbp]
  \centering
  \includegraphics[width=.92\columnwidth]{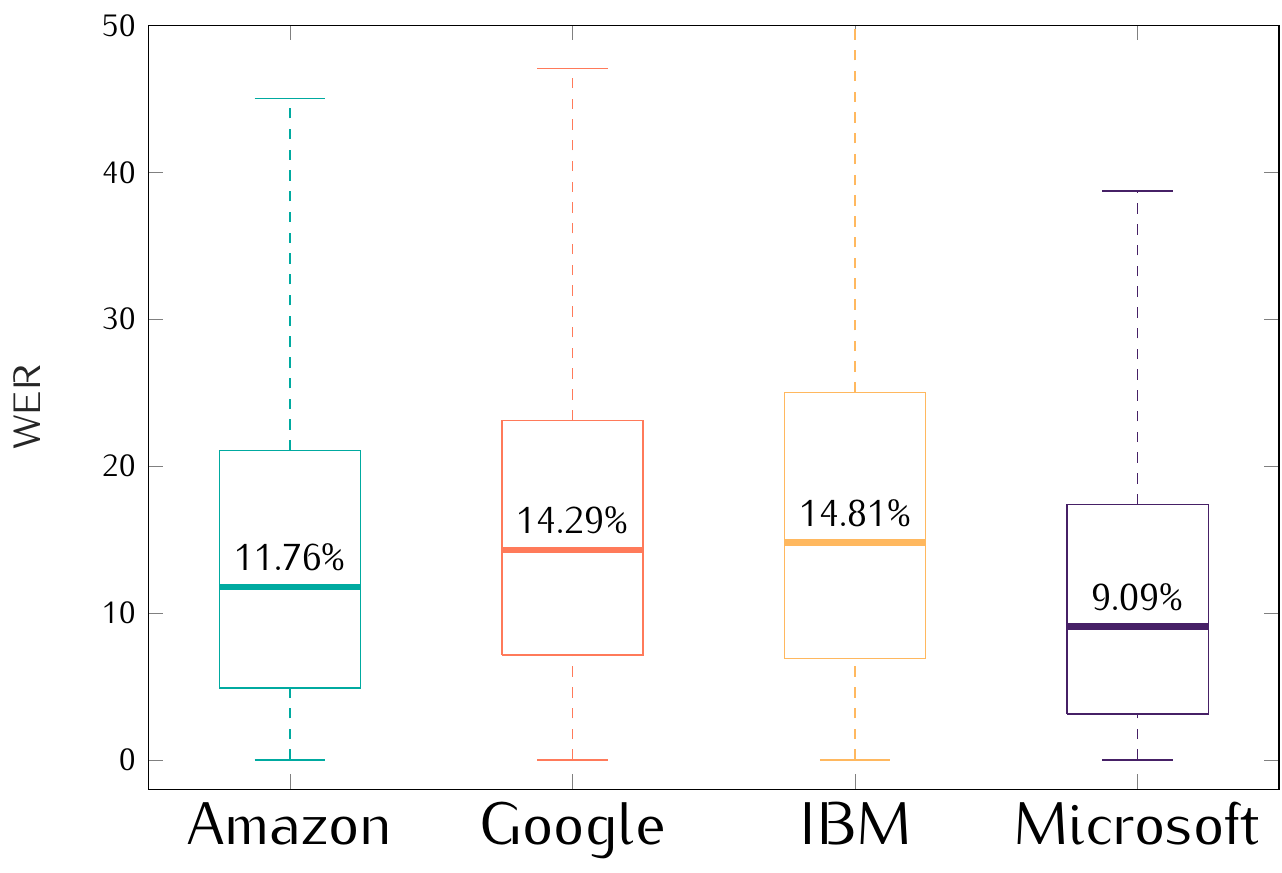}
    \begin{tabular}{|l|cccc|}
    \hline
          & {Amazon} & {Google} & {IBM} & {Microsoft} \\ \hline
    WER   & 11.76\% & 14.29\% & 14.81\% & \textbf{9.09\%} \\
    MER   & 11.54\% & 14.29\% & 14.29\% & \textbf{9.09\%} \\
    WIL   & 0.19  & 0.25  & 0.24  & \textbf{0.16} \\
    \hline
    \end{tabular}%
  \caption{Evaluation on clean audio. Upper, WER distributions; lower, median values. ({\scriptsize \sc{STTs accessed in February 2021}})}
  \label{tab:metric_clean}%
\end{table}%

\section{Results}

\subsection{Clean speech}

For clean speech, Microsoft Azure performed quite well, with a WER at $9.09\%$ which is close to the advertised rate. Amazon Transcribe took the second place with WER $11.76\%$. Google Cloud and IBM Waston gave similar WER ($14.29\%$ and $14.81\%$). 
These WER are actually very good already. 
According to the public DeepSpeech model \cite{Hannun2014Deep} from Mozila, trained with a mixed French dataset "CommonVoice + CssTen + LinguaLibre + Mailabs + Tatoeba + Voxforge", the WER on test dataset is $19.5\%$ (result retrieved on March 10\textsuperscript{th} 2021) \cite{Jaco-Assistant2021DeepSpeech}.
The gain with cloud STT API is between $24\% - 53\%$.

\subsection{Noisy speech}

After mixing five different levels of environmental noise, Microsoft Azure gave a quite good global WER $11.11\%$ (\Cref{tab:metric_noisy_all}). Amazon Transcribe and Google Cloud showed the same WER at $20\%$. But IBM Waston failed at certain point. Its global WER is $29.63\%$, with a word-information-lost rate at $43\%$ (0.43) which is unfortunately high. 

\begin{table}[!htbp]
  \centering
    \begin{tabular}{|l|cccc|}
    \hline
          & Amazon & Google & IBM   & Microsoft \\ \hline
    WER   & 20.00\% & 20.00\% & 29.63\% & \textbf{11.11\%} \\
    MER   & 19.64\% & 20.00\% & 28.57\% & \textbf{11.11\%} \\
    WIL   & 0.31  & 0.33  & 0.43  & \textbf{0.19} \\
    \hline
    \end{tabular}%
  \caption{Evaluation on all noisy audio (5 SNR levels combined), median values}
  \label{tab:metric_noisy_all}%
\end{table}%

\begin{figure}[!htb]
\includegraphics[width=\columnwidth]{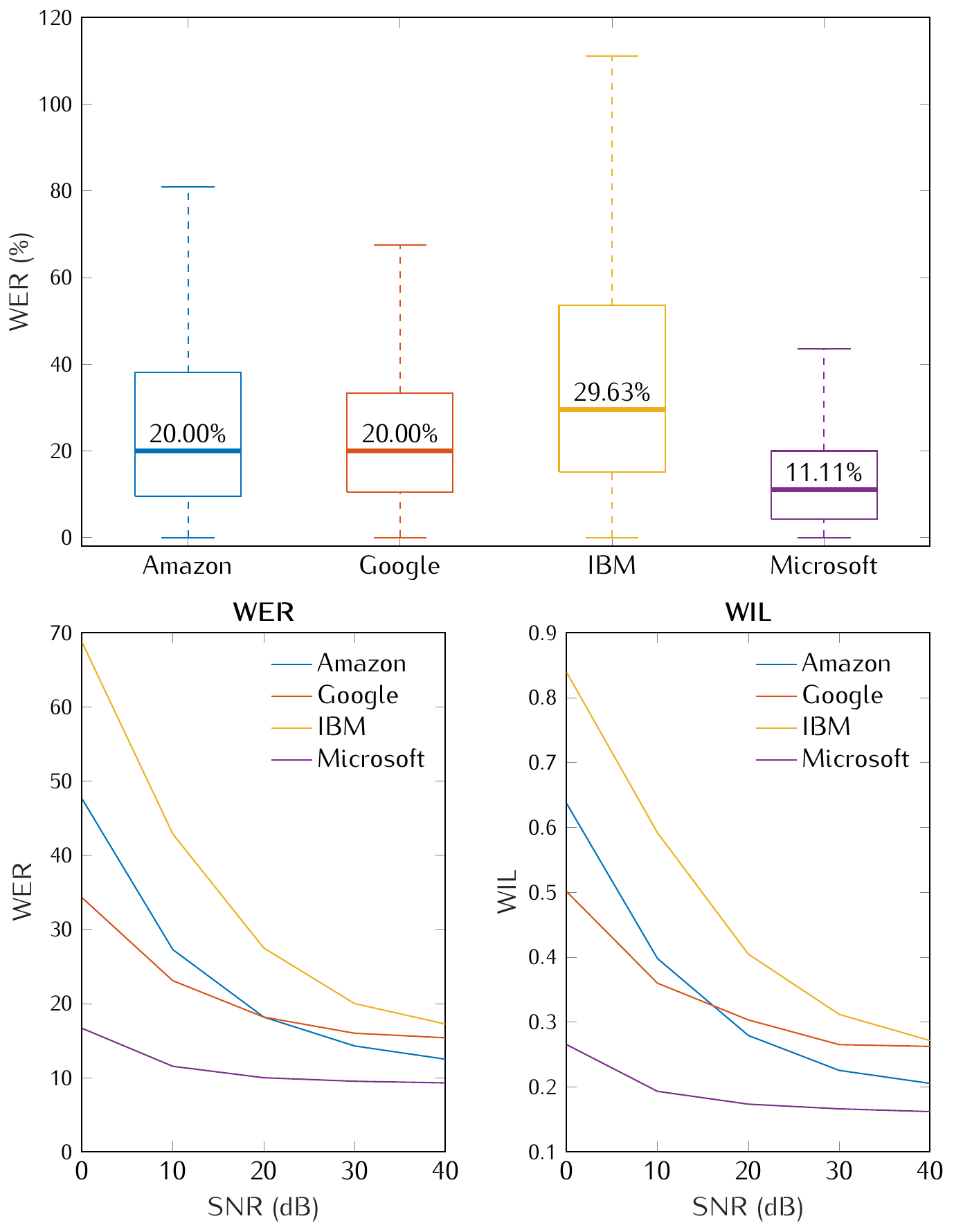}
\caption{Evaluation on mixed noisy speech by five signal-noise-ratio levels; upper, wer distributions; lower, median value for each level. ({\scriptsize \sc{STTs accessed in February 2021}})}
\label{fig:metric_noisy_snrdb}%
\end{figure}

At individual SNR level, as shown in \Cref{fig:metric_noisy_snrdb}, Microsoft Azure is the most robust to noise. The variation across different noise levels is quite small. 
In highly noisy environment, the WER from transcription by IBM Waston can be more than $100\%$. While other STTs would be at worst less than $50\%$. 

The exceptional STT performance of Microsoft Azure is due to that Microsoft has been working intensively on Artificial Intelligence based noise suppression. This environmental noise dataset MS-SNSD comes from Microsoft. 
The noise suppression should be already in the pipeline of their Speech-to-Text models. 
Actually, in December 2020 Microsoft introduced background noise suppression functionality in Microsoft Teams meetings \cite{Microsoft2020Reduce}. To achieve this, they used 760 hours of clean speech data and 180 hours of noise data. These data are now released for Interspeech 2021 Deep Noise Suppression Challenge \cite{Reddy2021Interspeech}. 

\begin{figure}[!htb]
\includegraphics[width=\columnwidth]{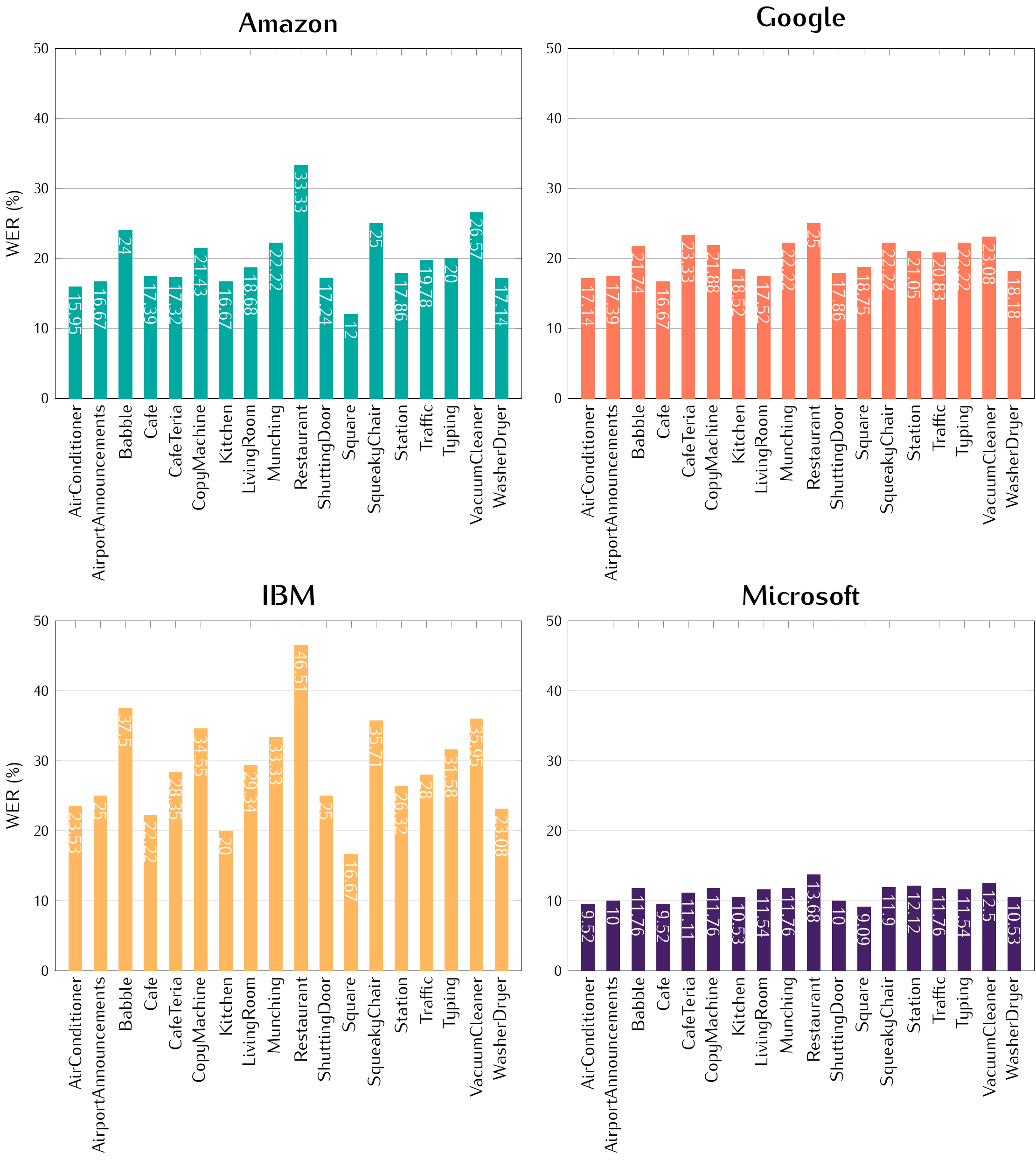}
\caption{WER by different noise types; median WER values from tests on all the five SNR noisy speech. ({\scriptsize \sc{STTs accessed in February 2021}})}
\label{fig:wer_noisetype}%
\end{figure}

The performance of STT depends also on the noise types. 
All the STT services are sensible to noise type \emph{Restaurant}. IBM Waston's WER reached $46.51\%$; Amazon Transcribe had also high WER for this type of noise. Google Cloud and Microsoft Azure dealt it better without shape WER changes. 
Background noise in environment \emph{Restaurant} could be a mixture of different noises (babble, conversation, munching, traffic etc.) which make it be more difficult for Speech-to-Text tasks. 
In general, Google Cloud and Microsoft Azure are more robust to environmental noise (variation and  standard deviation of the median WER are $6.5\%$ and $2.6\%$ for Google Cloud;  $1.4\%$, $1.2\%$ for Microsoft Azure); Amazon Transcribe can be placed in the second rank with $24\%$ and $4.9\%$. As for IBM Waston, as shown previously, it can fail in many cases when the background noises are too strong. It suffered also strong performance variation $53.7\%$ and  $7.3\%$ of standard deviation of the median WER.

\subsection{Main STT errors}

The main source of errors contributed to WER is the substitution. 
For clean speech, or less noisy speech, the percentage of substitution $S$ is generally much higher than deletion $D$ and insertion $I$. 
When speech becomes highly noisy (SNR lower than \SI{10}{\decibel}), deletion $D$ percentage increased much more. 
STT service from Microsoft Azure is quite robust to noisy environment, there's practically no change for SNR from \SI{40}{\decibel} to \SI{10}{\decibel}. Only in the tested case when mixing directly noise and speech \SI{0}{\decibel}, the deletion $D$ and substitution $S$ increased slightly. 
However, the changes are much more significant for other three STT services, especially for IBM Waston. 

\begin{figure}[!htb]
\centering
\includegraphics[width=0.95\columnwidth]{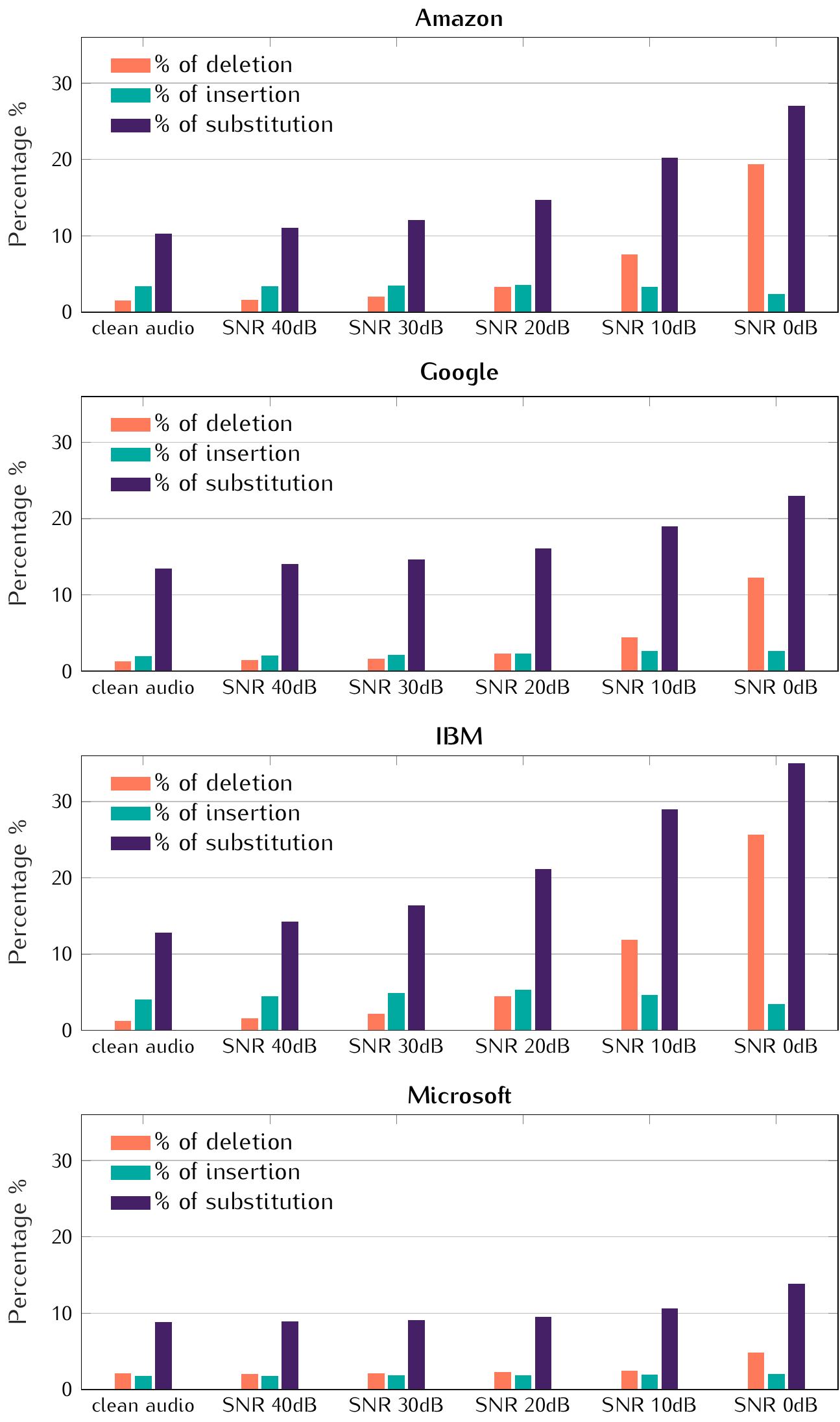}
\caption{Main transcription errors distribution (mean values. The median percentage values for lower SNR are zero, less meaningful for presentation)}
\label{fig:error_hdis_mean}%
\end{figure}

There's also inter-speakers difference of WER. Amount the 42 speakers, all the four STTs had more difficulty to transcribe speech from speaker {L23\_P08}. 

\begin{figure}[!htb]
\centering
\includegraphics[width=0.95\columnwidth]{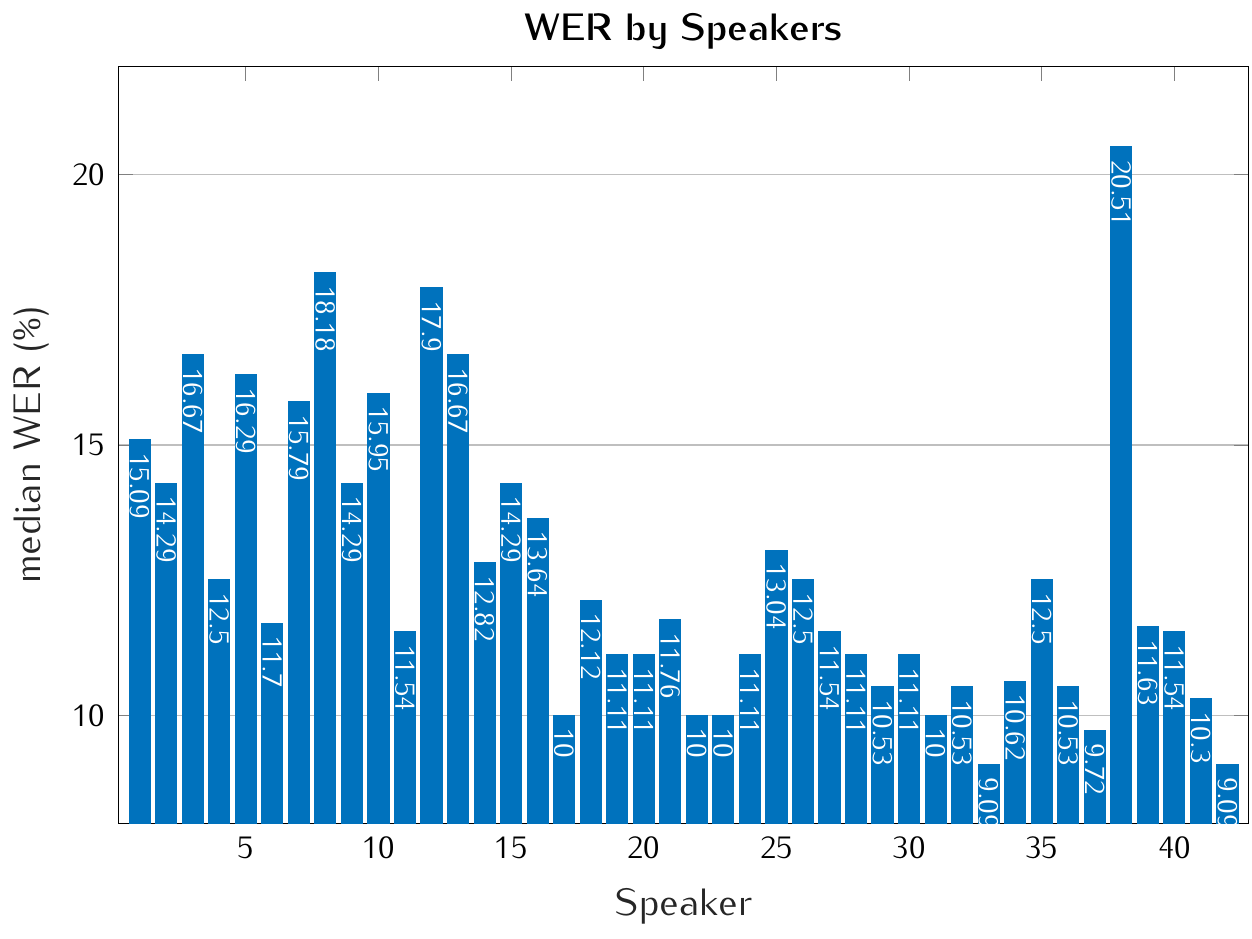}
\caption{Word Error Rate by speakers (median values) on clean speech for all the four STT systems.}
\label{fig:error_hdis}%
\end{figure}

\subsection{Transcription job time}

In a production application, the STT service must be as responsive as possible. 
Google Cloud is the fastest about the four tested APIs, with a median value at 1.76 second par job. 
Microsoft Azure is also fast, 3.51 second per transcription job. 
IBM Waston is slower and require 5.43 second to complete the job. 
It's not surprising that Amazon Transcribe is the slowest STT service, with 27 second per job. Some transcription jobs can take up to 200 second. Even it's possible to send up to 100 jobs in parallel, single job waiting is not acceptable for any real world application. This time requirement does not include the data transfer time to Amazon S3 storage: with upload speed 100-700 kbps, for a large amount of data, this can take already quite some time to complete. Though it's possible to call Amazon Transcribe for steaming usage, it's not convenient for non-real-time scenario. 

One of the potential reasons of the additional seconds from Google Cloud and IBM Waston, could be that Microsoft Azure's returns less complete transcription information than the other three. 

\begin{figure}[!htb]
\includegraphics[width=\columnwidth]{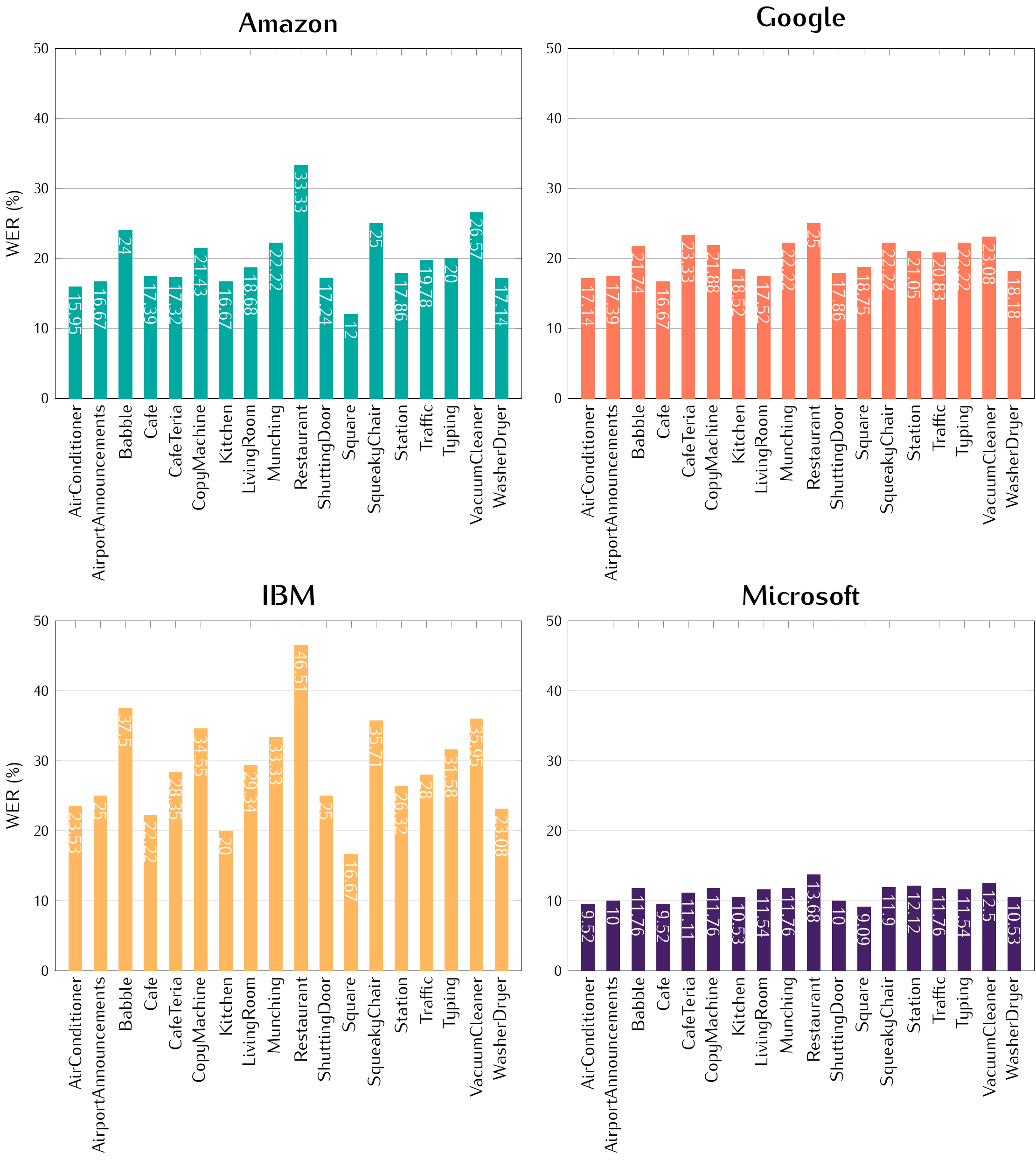}
\caption{Transcriptio job time in second for all the four STT systems}
\label{fig:jobtime}%
\end{figure}

Another observation is the server responsiveness: job completion time with Microsoft Azure is almost linear to the audio duration. The variation is also very tight. But for other APIs, though the relationship could be regarded as linear, the variation is much larger. Speeches with same length would require 2 to 4 times more execution time to complete the task. 

\section{Discussion}

In this work, we evaluated the four most used Speech-to-Text API on French speech from four Computing Cloud: Amazon Transcribe, Google Cloud, IBM Waston and Microsoft Azure. 5 levels of different environment noises are mixed with 6690 clean speeches (17 hours). 100 hours speech tests per STT API gave 400 hours speech transcription. 

The results showed that Microsoft Azure's STT service provided the lowest Word Error Rate (median 9\%). It's also very robust to common environment noise, even in strong noise environment, the median WERs are only around 16.67\%. 
STT from Amazon Transcribe and Google Cloud performed well, their WER are respectively at 11.76\% and 14\%. Amazon Transcribe works better in relatively quiet environment while Google Cloud is better for noisy speech. 
IBM Waston's STT service can provide reasonable results with a median WER at 14.29\%. But when the speech is recorded in noisy environment, the WER can go up to around 70\% which is difficult to be used. 
In general, when the signal-to-noise ratio is higher than \SI{20}{\decibel}, the WERs are still acceptable. However, if SNR drops lower than \SI{20}{\decibel}, except Microsoft Azure, all the three APIs will have difficulties to recognize correctly the speech. 
Among the 18 environment noise types, Restaurant type is the most difficult one to deal with for all the four STT APIs. 

When the work is time-constraint, Google Cloud will be the first choice with fastest response time and a reasonable word error rate. Amazon Transcribe can be used when the framework of the project is on the platform of Amazon Web Services. The parallel job can help to reduce the total transcription time, however, per job time is too longer than any other STT service. In average, one transcription job on Amazon Transcribe is 15 times longer than the same job on Google Cloud. 
Otherwise, the general suggestion will be Microsoft Azure, lowest WER and high robustness to noise. It's more suitable for precision-constraint applications. 

\Urlmuskip=0mu plus 1mu\relax
\renewcommand\refname{References}
\bibliographystyle{IEEEtran}  
\small{\bibliography{bib}}

\begin{thebibliography}{10}
\providecommand{\url}[1]{#1}
\csname url@rmstyle\endcsname
\providecommand{\newblock}{\relax}
\providecommand{\bibinfo}[2]{#2}
\providecommand\BIBentrySTDinterwordspacing{\spaceskip=0pt\relax}
\providecommand\BIBentryALTinterwordstretchfactor{4}
\providecommand\BIBentryALTinterwordspacing{\spaceskip=\fontdimen2\font plus
\BIBentryALTinterwordstretchfactor\fontdimen3\font minus
  \fontdimen4\font\relax}
\providecommand\BIBforeignlanguage[2]{{%
\expandafter\ifx\csname l@#1\endcsname\relax
\typeout{** WARNING: IEEEtran.bst: No hyphenation pattern has been}%
\typeout{** loaded for the language `#1'. Using the pattern for}%
\typeout{** the default language instead.}%
\else
\language=\csname l@#1\endcsname
\fi
#2}}

\bibitem{Saon2017English}
\BIBentryALTinterwordspacing
G.~Saon, G.~Kurata, T.~Sercu, K.~Audhkhasi, S.~Thomas, D.~Dimitriadis, X.~Cui,
  B.~Ramabhadran, M.~Picheny, L.-L. Lim, \emph{et~al.}, ``English
  conversational telephone speech recognition by humans and machines,''
  \emph{arXiv preprint arXiv:1703.02136}, 2017. [Online]. Available:
  \url{https://arxiv.org/abs/1703.02136}
\BIBentrySTDinterwordspacing

\bibitem{Picovoice2020Speech}
\BIBentryALTinterwordspacing
Picovoice, ``Speech-to-text benchmark,'' \emph{GitHub}, 2020. [Online].
  Available: \url{https://github.com/Picovoice/speech-to-text-benchmark}
\BIBentrySTDinterwordspacing

\bibitem{Chiu2018State}
C.~{Chiu}, T.~N. {Sainath}, Y.~{Wu}, R.~{Prabhavalkar}, P.~{Nguyen}, Z.~{Chen},
  A.~{Kannan}, R.~J. {Weiss}, K.~{Rao}, E.~{Gonina}, N.~{Jaitly}, B.~{Li},
  J.~{Chorowski}, and M.~{Bacchiani}, ``State-of-the-art speech recognition
  with sequence-to-sequence models,'' in \emph{2018 IEEE International
  Conference on Acoustics, Speech and Signal Processing (ICASSP)}, April 2018,
  pp. 4774--4778.

\bibitem{Saon2015IBM}
\BIBentryALTinterwordspacing
G.~Saon, H.-K.~J. Kuo, S.~Rennie, and M.~Picheny, ``The ibm 2015 english
  conversational telephone speech recognition system,'' \emph{arXiv preprint
  arXiv:1505.05899}, 2015. [Online]. Available:
  \url{https://arxiv.org/abs/1505.05899}
\BIBentrySTDinterwordspacing

\bibitem{Xiong2018Microsoft}
W.~{Xiong}, L.~{Wu}, F.~{Alleva}, J.~{Droppo}, X.~{Huang}, and A.~{Stolcke},
  ``The microsoft 2017 conversational speech recognition system,'' in
  \emph{2018 IEEE International Conference on Acoustics, Speech and Signal
  Processing (ICASSP)}, April 2018, pp. 5934--5938.

\bibitem{Besacier2014Word}
L.~Besacier, B.~Lecouteux, N.-Q. Luong, K.~Hour, and M.~Hadj~Salah, ``Word
  confidence estimation for speech translation,'' in \emph{International
  Workshop on Spoken Language Translation}, Lake Tahoe, United States, Dec.
  2014.

\bibitem{Le2016Joint}
N.-T. Le, B.~Lecouteux, and L.~Besacier, ``Joint asr and mt features for
  quality estimation in spoken language translation,'' in \emph{International
  Workshop on Spoken Language Translation}, Seattle, United States, Dec. 2016.

\bibitem{Reddy2019Scalable}
C.~K. Reddy, E.~Beyrami, J.~Pool, R.~Cutler, S.~Srinivasan, and J.~Gehrke, ``A
  scalable noisy speech dataset and online subjective test framework,'' in
  \emph{Proc. Interspeech 2019}, 2019, pp. 1816--1820.

\bibitem{Morris2004From}
A.~C. Morris, V.~Maier, and P.~Green, ``From wer and ril to mer and wil:
  improved evaluation measures for connected speech recognition,'' in
  \emph{Eighth International Conference on Spoken Language Processing}, 2004.

\bibitem{Hannun2014Deep}
A.~Y. Hannun, C.~Case, J.~Casper, B.~Catanzaro, G.~Diamos, E.~Elsen,
  R.~Prenger, S.~Satheesh, S.~Sengupta, A.~Coates, and A.~Y. Ng, ``Deep speech:
  Scaling up end-to-end speech recognition,'' \emph{CoRR}, vol. abs/1412.5567,
  2014.

\bibitem{Jaco-Assistant2021DeepSpeech}
\BIBentryALTinterwordspacing
Jaco-Assistant, ``Deepspeech-polyglot,'' \emph{GitLab}, 2021. [Online].
  Available: \url{https://gitlab.com/Jaco-Assistant/deepspeech-polyglot}
\BIBentrySTDinterwordspacing

\bibitem{Microsoft2020Reduce}
\BIBentryALTinterwordspacing
Microsoft, ``Reduce background noise in microsoft teams meetings with ai-based
  noise suppression,'' 2020. [Online]. Available:
  \url{https://techcommunity.microsoft.com/t5/microsoft-teams-blog/reduce-background-noise-in-microsoft-teams-meetings-with-ai/ba-p/1992318}
\BIBentrySTDinterwordspacing

\bibitem{Reddy2021Interspeech}
\BIBentryALTinterwordspacing
C.~K. Reddy, H.~Dubey, K.~Koishida, A.~Nair, V.~Gopal, R.~Cutler, S.~Braun,
  H.~Gamper, R.~Aichner, and S.~Srinivasan, ``Interspeech 2021 deep noise
  suppression challenge,'' \emph{arXiv preprint arXiv:2101.01902}, 2021.
  [Online]. Available: \url{https://arxiv.org/abs/2101.01902}
\BIBentrySTDinterwordspacing

\end{thebibliography}

\end{document}